\documentclass[conference]{IEEEtran}
\IEEEoverridecommandlockouts
\usepackage{cite}
\usepackage{amsmath,amssymb,amsfonts}
\usepackage{algorithmic}
\usepackage{graphicx}
\usepackage{textcomp}
\usepackage{xcolor}
\usepackage{comment}
\usepackage{booktabs}
\usepackage{multirow, makecell}
\usepackage{multicol}
\ifCLASSOPTIONcompsoc
\usepackage[caption=false,font=normalsize,labelfon
t=sf,textfont=sf]{subfig}
\else
\usepackage[caption=false,font=footnotesize]{subfi
g}
\fi
\def\BibTeX{{\rm B\kern-.05em{\sc i\kern-.025em b}\kern-.08em
    T\kern-.1667em\lower.7ex\hbox{E}\kern-.125emX}}
\begin{document}

\title{Improving Taxonomic Image-based Out-of-distribution Detection With DNA Barcodes\thanks{The work was funded by Research Council of Finland project 333497.}}

\begin{comment}
\author{\IEEEauthorblockN{Mikko Impiö}
\IEEEauthorblockA{\textit{Finnish Environment Institute Syke} \\
Helsinki, Finland \\
email address or ORCID}
\and
\IEEEauthorblockN{2\textsuperscript{nd} Given Name Surname}
\IEEEauthorblockA{\textit{dept. name of organization (of Aff.)} \\
\textit{name of organization (of Aff.)}\\
City, Country \\
email address or ORCID}
}

\end{comment}

\author{\IEEEauthorblockN{Mikko Impiö\IEEEauthorrefmark{1} and
Jenni Raitoharju\IEEEauthorrefmark{1}\IEEEauthorrefmark{2}
}
\IEEEauthorblockA{\IEEEauthorrefmark{1}Finnish Environment Institute, Quality of Information, Finland}
\IEEEauthorblockA{\IEEEauthorrefmark{2}University of Jyväskylä, Faculty of Information Technology, Finland}
}

\maketitle

\begin{abstract}

Image-based species identification could help scaling biodiversity monitoring to a global scale.
Many challenges still need to be solved in order to implement these systems in real-world applications.
A reliable image-based monitoring system must detect out-of-distribution (OOD) classes it has not been presented before.
This is challenging especially with fine-grained classes.
Emerging environmental monitoring techniques, DNA metabarcoding and eDNA, can help by providing information on OOD classes that are present in a sample.
In this paper, we study if DNA barcodes can also support in finding the outlier images based on the outlier DNA sequence's similarity to the seen classes.
We propose a re-ordering approach that can be easily applied on any pre-trained models and existing OOD detection methods.
We experimentally show that the proposed approach improves taxonomic OOD detection compared to all common baselines.
We also show that the method works thanks to a correlation between visual similarity and DNA barcode proximity.
The code and data are available at https://github.com/mikkoim/dnaimg-ood.

\end{abstract}


\section{Introduction}

The state of the natural environment has been significantly altered by humans and the increase of species extinction has prompted for action in environmental monitoring \cite{ipbes2019Summary}.
Efforts are currently being scaled up to address the challenge of monitoring the entirety of our planet.
Automated methods based on computer vision and imaging devices that can collect more data from the environment and species are gaining a lot of interest among ecologists \cite{hoye2021Deep,wuhrl2024Entomoscope, deschaetzen2023Riverine}.

With automatic monitoring methods gaining interest, it is important to consider the trustworthiness of these systems.
Image classification methods based on Deep Neural Networks (DNNs) have been shown to be effective when the models are evaluated with the same data distribution they were trained on \cite{schneider2022Bulk, bjerge2023Accurate}.
However, in a realistic use case the trained model is presented to the "open world", where samples from unseen classes can be presented to them.
This is known as the "known unknown" domain of pattern recognition, which has been extensively studied via out-of-distribution (OOD) problems \cite{zhang2020Robust}.
It is well known, that DNNs can output high confidence outputs for out-of-distribution classes and even for inputs containing only noise \cite{nguyen2015Deep}.
OOD detection is especially relevant with arthropod and insect classification, as a vast amount of these groups remain undescribed \cite{stork2018How}.
Confidently made incorrect classifications can have undesired effects on the biological indices calculated from taxa lists, as well as on biomass and abundance estimates.

Molecular methods such as deoxyribonucleic acid (DNA) barcoding \cite{hebert2005Promise} have been proposed as an answer to the accurate identification problem.
The amount of sequencing data and availability of reference sequence libraries, such as the Barcode of Life database (BOLD) \cite{ratnasingham2007bold}, are increasing as sequencing technologies are becoming affordable \cite{srivathsan2021ONTbarcoder}.
Molecular methods can identify the species, but are not accurate with biomass or abundance estimation \cite{creedy2019accurate}.
Thus, manual classification or automatic recognition based on images remain the most viable options for these tasks.

OOD detection has been studied broadly for different modalities, including images, text, and genomic data \cite{hendrycks2022Scaling, liu2020Energybased, ren2019Likelihood}.
Methods are typically based on ranking the samples using an \textit{OOD scoring method} that produces a metric score.
A high metric score is then associated with the likelihood of a sample being an outlier.

Taxonomic out-of-distribution detection has usually been done with benchmark datasets.
These benchmark datasets often have classes that are easily distinguishable from each other, with only few fine-grained classes, such as dog breeds or birds.
Hendrycs et al. \cite{hendrycks2022Scaling} recognize this problem and observe that with easily distinguishable classes the outputs are more concentrated, while with fine-grained classes outputs are divided between the highly similar classes.
This problem is also present with finely-grained insect datasets, making them a good example of challenging OOD detection.

This paper proposes using DNA barcoding data in image-based OOD detection in fine-grained taxa identification, which is a new research area.
OOD detection purely for genetic data has been studied before \cite{bai2022MLROOD}. 
In zero-shot-classification, using DNA side information for improving accuracy was proposed by \cite{badirli2023Classifying}.
Using side information in OOD detection has not been as common as in other areas of study.
We show in this paper that using available DNA side information can produce better OOD detection performance.

\begin{figure*}[htpb]
    \centering
    \includegraphics[width=0.95\linewidth]{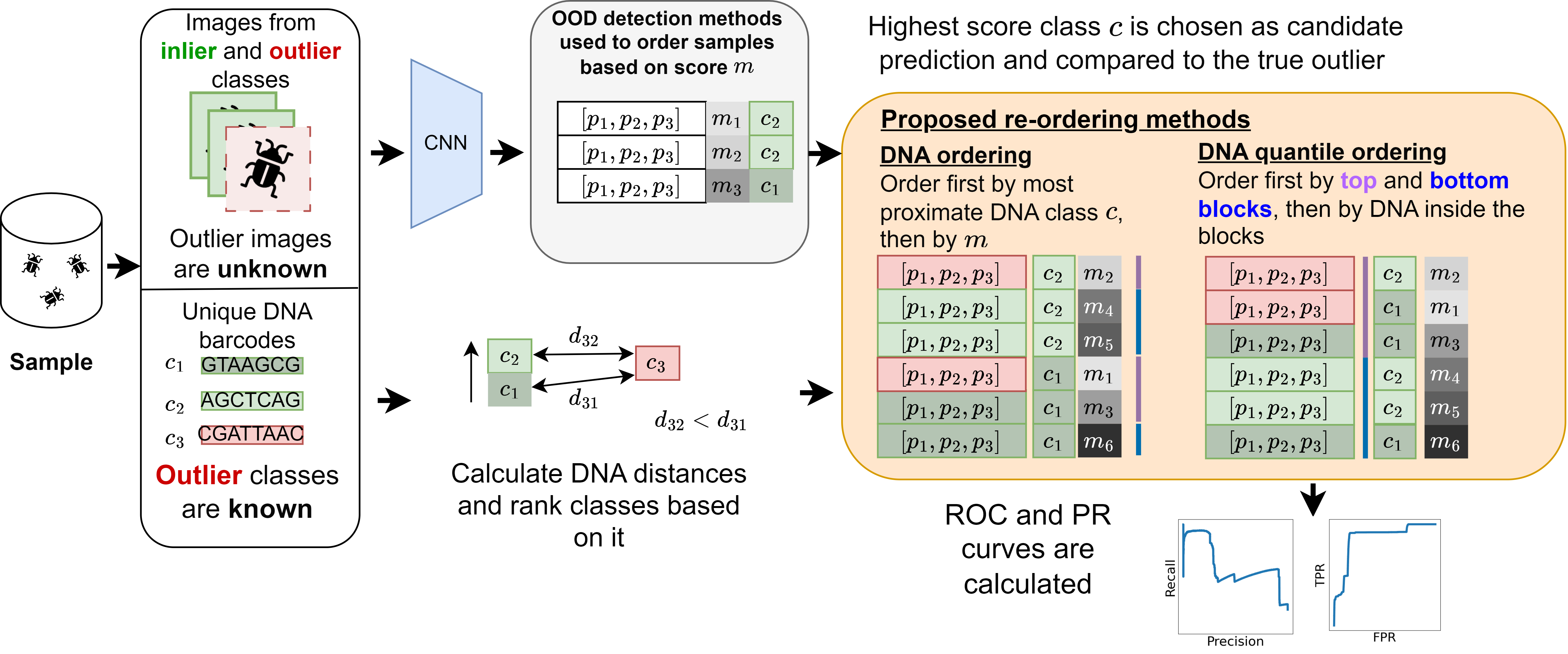}
    \caption{Overview of the proposed OOD detection approach}
    \label{fig:overview}
\end{figure*}

\section{Proposed method}

We propose an approach for outlier (unseen taxa) detection in taxonomic identification using DNA barcodes as side information.
The proposed method serves as an additional re-ordering component applied after a more commonly used scoring method, such as Energy \cite{liu2020Energybased} or MaxLogit \cite{hendrycks2022Scaling}.
Our assumption is that using DNA barcode proximity for re-ordering the OOD detection scoring outputs can improve outlier detection.
This is based on the assumption that DNA barcodes encode some aspects of visual similarity.

The practical use case for the method could be the following: We have a sample of specimens that we want to image using an imaging device and classify with a deep neural network (DNN).
The classifier has been previously trained with a set of $C$ inlier classes, $\mathcal{C}_{in}$.
We can also extract DNA barcodes from the sample using DNA metabarcoding or eDNA approaches \cite{pawlowski2022environmental}.
This means that we can sequence the specimens in bulk and get a list of the taxa (classes) present in the collection, but the DNA sequences can not be assigned to specific specimens (images).
We assume that the DNA list reveals the presence of a set of $C_{out}$ outlier classes, $\mathcal{C}_{out}$, we have not encountered when training the classifier.
We also assume that the classes in $\mathcal{C}_{out}$ can be found in DNA barcode reference databases.
Thus, the DNA list reveals information on the presence of outlier classes, but we cannot know which images/specimens they correspond to.
We want to find out the most probable outliers in the sample.
In actual monitoring situations, the candidate images could be then verified by a human taxonomist.

The proposed method works in four phases: 1) Classification, 2) OOD scoring, 3) DNA distance ranking, and 4) Re-ordering the samples based on DNA distances. Finally, metrics are calculated for the re-ranked samples. An overview of the process can be seen in Fig.~\ref{fig:overview}.

\subsection{Classification}
For classification, any commonly used deep neural network model can be used. The model $f$ has parameters $\theta$, it takes an input image $x$ and outputs a discrete logit score vector $f_{\theta}(x) = \mathbf{s} = [s_0, s_1, ..., s_{C-1}]$ of length $C$, where $C$ is the number of classes the model was trained on. A logit score $s_i$ can be transformed into a softmax probability score $p_i$ by
\begin{equation}
p_i = \frac{e^{s_i}}{\sum^{C-1}_{j=0}e^{s_j}}.
\end{equation}
The softmax probability score represents the heuristic probability of the image to belong to this class, such that $\sum^{C-1}_{i=0} p_i = 1$.
The most likely class assignment for an image can be calculated by taking the class index of the highest score $c = \text{argmax}(\mathbf{s})$.
This predicted class is later used in re-ordering.

\subsection{Out-of-distribution scoring}

Previous OOD detection methods calculate a OOD scoring metric $m$ from the DNN output $\mathbf{s}$. This can be used to rank the samples in a decreasing order based on their probability to be outliers.
This order also serves as a baseline for the DNA re-ordering performed later.
From the ranking, we can choose the first $p$ percent items as the most likely outlier candidates.
The percentage is an arbitrary threshold and can be chosen freely, with the trade-off between true and false positive rates.
Lower thresholds ensure that all inlier classes are classified as inliers with the cost of more outliers being assumed as inliers.
Higher thresholds ensure that more outliers are correctly detected with the cost of more inliers being assumed as outliers.

We use different OOD scoring methods proposed in the literature and others designed for the study that we could not easily find in the literature: 1) Maximum Sigmoid Probability (MSP) \cite{hendrycks2017Baseline}, 2) Maximum Logit value (MaxLogit) \cite{hendrycks2022Scaling}, 3) Energy score\cite{liu2020Energybased}, 4) Entropy of the sigmoid distribution, 5) Ratio between second highest and highest logit, and 6) Ratio between second highest and highest sigmoid probability.
The MSP is simply the maximum value from the sigmoid probability output, $m_{MSP} = \max(\mathbf{p})$, and MaxLogit refers to the maximum value of the raw logit output $\mathbf{s}$.
In practice, to get a decreasing order, the values are multiplied by $-1$.
The energy function is also calculated from the logit output, by $m_{\text{energy}} = -\log \left( \sum^{C-1}_{i=0} e^{s_i} \right)$ \cite{liu2020Energybased}.
Entropy is calculated from the sigmoid probabilities with $m_{\text{entropy}} = \sum^{C-1}_{i=0} p_i \log(p_i)$.
The ratio scores are simply either $\text{max2}(\mathbf{y}) / \max(\mathbf{y})$, where $\mathbf{y}$ is either $\mathbf{s}$ (logit) or $\mathbf{p}$ (sigmoid). The function $\text{max2}(\mathbf{y})$ returns the second highest value of $\mathbf{y}$.

\subsection{DNA distances}
\label{ssec:dnadistance}
We assume that the bulk DNA sample can be sequenced and a list of present taxa $\mathcal{C}_{DNA}$ can be attained by comparing the sequences to a reference database, such as the Barcode of Life Database (BOLD) \cite{ratnasingham2007bold}.
Each taxa class has a representative DNA sequence consisting of nucleotide bases $\mathbf{b}\in \{A,G,T,C\}^{n}$, which in practice is a DNA barcode \cite{hebert2003Biological} of length $n$.
We can calculate a distance $d_{ab} = dist(\mathbf{b}_a, \mathbf{b}_b)$ between two sequences.
Here, $dist(\cdot, \cdot)$ is a DNA distance function.

There are several possible ways of calculating distances between DNA sequences.
The most simple one is the percentage of sites that differ between sequences, referred to as the "raw" difference.
In evolutionary biology, the Kimura 2-parameter distance (referred to as "K80") \cite{kimura1980simple} is more commonly used.
It assumes that the probabilities for transition and transversion substitutions are different.
Transition means that the substitution happens inside a pyrimidine (C,T) or purine (A,G) pairs, while transversion refers to the substitution between and pyrimidines and purines. We experimented with both raw and K80 DNA similarities.

The outlier class set can be found from the set of present taxa by $\mathcal{C}_{out} = \mathcal{C}_{\text{DNA}} \setminus \mathcal{C}_{in}$.
In this paper, we consider the special case of $|\mathcal{C}_{out}| = 1$. As we have a single outlier class $c_{out}$, we can calculate the DNA distance to all inlier classes so that we get a list of distances $[d_0, d_1, d_2, d_{C-1}]$ for all outlier-inlier pairs.
This list is then ranked, so that the inlier class with the shortest distance to the outlier is first.

\subsection{Re-ordering based on DNA distances}

Our main contribution is a re-ordering method that serves as an additional component after OOD detection has been done for the DNN outputs.
We use the DNA distance ranking as side information to re-order the initial OOD ranking done from the images.
We propose two different approaches: 1) DNA ordering and 2) DNA quantile ordering.

In both approaches, images are prioritized based on the DNA-based similarity (Section~\ref{ssec:dnadistance}) between their predicted class and the outlier class.
In the first re-ordering approach, all images classified as the inlier class with the highest DNA similarity to the outlier class are taken and then arranged based on their OOD scoring metric.
Next, images from the class with the second highest DNA similarity  are taken, and so on, following the DNA similarity ranking.
Thus, samples classified by the DNN as the inlier class closest to the known outlier, determined by DNA proximity, are considered the most probable specimens belonging to the outlier class in the order of their OOD scoring.

The DNA quantile method adds an additional step of first considering the block of $q$ percent of the specimens with the highest OOD score and re-ordering them by DNA as described above.
Next, the rest $1-q$ percent of the specimens are ordered.
This approach takes into account that the OOD scoring metric might initially give us better information about a specimen being an outlier, and the classification along with the DNA-based class similarity is considered second.
Finally, the order inside each classification group is determined by the OOD scoring metric as described.

\section{Experiments}

\subsection{Experimental setup}

\subsubsection{Dataset}
We use the FinBenthic 2 dataset\footnote{https://etsin.fairdata.fi/dataset/a11cdc26-b9d0-4af1-9285-803d65a696a3}\cite{arje2020Human} for all of our experiments.
It has 460 009 images from 9631 specimens, belonging to 39 different taxa.
26 of the taxa are classified to the taxonomic level of a species, while the rest are classified to genus level, except one class, \textit{Simuliidae}, which is on family level.
Each specimen contains a maximum number of 50 images.
Each image is classified separately.
The largest class, \textit{Simuliidae}, contains 44240 images from 887 specimens, while the smallest class, \textit{Hydropsyche saxonica}, has 490 images from 17 specimens.

As there is no DNA data corresponding to the specimens in the FinBenthic 2 dataset, we collected a toy dataset representing image-DNA pairs.
For each 39 classes, we collected a DNA sequence representing this taxon.
The DNA data was collected from BOLD \cite{ratnasingham2007bold}.
The samples were chosen by searching the database for sequences from a taxa and choosing a random entry.
For the classes corresponding to taxa higher than species level, we chose a random species in the corresponding genus or family.
Only high-quality sequences from the COI gene were considered.
Samples were chosen to be from Finland to match with the image dataset origin.
After collecting the DNA dataset, we aligned the sequences using the MUSCLE multiple sequence alignment method \cite{edgar2004MUSCLE}.
The DNA sequence dataset is available at \textit{https://github.com/mikkoim/dnaimg-ood}.

\subsubsection{Classification setup}

As the dataset has 39 different classes, we trained a total of 39 different models.
Each model was trained with an inlier set of 38 classes, with one taxon serving as the outlier.
The dataset was split to train and test sets so that $80\%$ of the inlier class specimens are in the training set, and all specimens from the outlier class and $20\%$ of specimens from each of the 38 inlier classes are in the test set.
The splits were grouped by specimen, so that all images of a single specimen belong either to the test or train split.

Models were trained for 30 epochs, using the AdamW \cite{loshchilov2019Decoupled} optimization algorithm.
As the DNN backbone we used the EfficientNet-b0 \cite{tan2019EfficientNet} architecture.
We used a batch size of 256, with a learning rate determined separately for each model with a learning rate search from the PyTorch Lightning library \cite{Falcon_PyTorch_Lightning_2019}.
The learning rate was in the interval $[2.291\cdot10^{-3}, 5.754\cdot10^{-3}]$ for all models.
The images were resized to 224x224 pixels, and random data augmentation was applied.
We used an assortment of data augmentation steps, including horizontal and vertical flips, geometric augmentations, color shifts, and pixel dropout.
Augmentation was not used with the test dataset.

\subsubsection{Implementation environment}

We used R for the DNA distance calculations, and the \textit{taxonomist} Python library for training the models.

\subsubsection{Evaluation metrics}

For the final evaluation of the DNA re-ordering methods, we use metrics commonly used in OOD detection literature: AUROC, AUPRC and FPR@95 metrics.
The AUROC and AUPRC are the areas under the Reciever Operating Characteristics and Precision-Recall curves.
The metrics correspond to calculating true positive rates (TPR), false positive rates (FPR), precision, and recall for all possible thresholds, and integrating the values as the area under the curves.
FPR@95 is calculated from the same AUROC curve, and corresponds to the FPR at the position where TPR is 95\%.

\subsection{Experimental results}

\subsubsection{Re-ordering based on DNA}

We calculated the AUPRC, AUROC, and FPR@95 metrics for all classes with all OOD scoring methods and DNA re-ordering methods with two different DNA distance metrics, raw and K80.
We observed that the raw and K80 DNA distances produced very similar outputs, often producing the exact same ranking and, thus, resulting in the same metrics.
The K80 was slightly better so we use it when reporting the results.
In Table \ref{tab:results}, we report the averaged metrics (using the K80 distance) with their standard deviations over all 39 classes.

\begin{table}[tbp]
    \caption{Re-ordering methods compared to OOD baseline order for different OOD scoring methods. Values are averages over 39 models trained for 39 different outlier classes. For DNA distance re-ordering, the K80 distance was used. $q=0.4$ was used in DNA quantile method.}
    \centering
    \resizebox{1\linewidth}{!}{%
    \begin{tabular}{llccc}
     \textbf{OOD}& \textbf{Re-ordering} & \multirow[m]{2}{*}{\textbf{AUPRC} $\uparrow$ }& \multirow[m]{2}{*}{\textbf{AUROC} $\uparrow$} & \multirow[m]{2}{*}{\textbf{FPR@95} $\downarrow$} \\
    \textbf{scoring} & \textbf{method }&  &  & \\
    \cline{1-5}
    \multirow[m]{3}{*}{Entropy} & OOD baseline & 0.317 ($\pm$0.22) & 0.808 ($\pm$0.10) & 0.501 ($\pm$0.17) \\
     & DNA & 0.444 ($\pm$0.26) & 0.806 ($\pm$0.18) & 0.467 ($\pm$0.30) \\
     & DNA quantile & \textbf{0.469 ($\pm$0.24)} & \textbf{0.864 ($\pm$0.06)} & \textbf{0.390 ($\pm$0.13)} \\
    \cline{1-5}
    \multirow[m]{3}{*}{MSP} & OOD baseline & 0.302 ($\pm$0.21) & 0.804 ($\pm$0.09) & 0.503 ($\pm$0.17) \\
     & DNA & 0.443 ($\pm$0.26) & 0.806 ($\pm$0.18) & 0.467 ($\pm$0.30) \\
     & DNA quantile & \textbf{0.468 ($\pm$0.24)} & \textbf{0.863 ($\pm$0.06)} & \textbf{0.395 ($\pm$0.14)} \\
    \cline{1-5}
    \multirow[m]{3}{*}{MaxLogit} & OOD baseline & 0.319 ($\pm$0.24) & 0.801 ($\pm$0.13) & 0.493 ($\pm$0.24) \\
     & DNA & 0.447 ($\pm$0.26) & 0.806 ($\pm$0.18) & 0.468 ($\pm$0.30) \\
     & DNA quantile & \textbf{0.455 ($\pm$0.22)} & \textbf{0.856 ($\pm$0.06)} & \textbf{0.337 ($\pm$0.08)} \\
    \cline{1-5}
    \multirow[m]{3}{*}{Energy} & OOD baseline & 0.317 ($\pm$0.23) & 0.800 ($\pm$0.13) & 0.493 ($\pm$0.24) \\
     & DNA & 0.447 ($\pm$0.26) & 0.806 ($\pm$0.18) & 0.468 ($\pm$0.30) \\
     & DNA quantile & \textbf{0.454 ($\pm$0.22)} & \textbf{0.855 ($\pm$0.06)} & \textbf{0.335 ($\pm$0.09)} \\
    \cline{1-5}
    \multirowcell{3}[0pt][l]{Ratio \\ (logit)} & OOD baseline & 0.253 ($\pm$0.17) & 0.762 ($\pm$0.10) & 0.605 ($\pm$0.13) \\
     & DNA & 0.435 ($\pm$0.26) & 0.804 ($\pm$0.18) & 0.469 ($\pm$0.30) \\
     & DNA quantile & \textbf{0.436 ($\pm$0.24)} & \textbf{0.833 ($\pm$0.08)} & \textbf{0.490 ($\pm$0.17)} \\
    \cline{1-5}
    \multirowcell{3}[0pt][l]{Ratio \\ (softmax)} & OOD baseline & 0.282 ($\pm$0.19) & 0.798 ($\pm$0.09) & 0.511 ($\pm$0.16) \\
     & DNA & 0.442 ($\pm$0.26) & 0.806 ($\pm$0.18) & 0.467 ($\pm$0.30) \\
     & DNA quantile & \textbf{0.464 ($\pm$0.24)} & \textbf{0.861 ($\pm$0.07)} & \textbf{0.408 ($\pm$0.14)} \\
    \cline{1-5}
    \end{tabular}}%
    \label{tab:results}%
\end{table}%

Our proposed DNA-based approach produces almost always better results than the OOD baseline.
The improvement between the baseline and plain DNA ordering is very slight or negligible, but the DNA quantile method is consistently better.

\begin{figure}
    \centering
    \includegraphics[width=\linewidth]{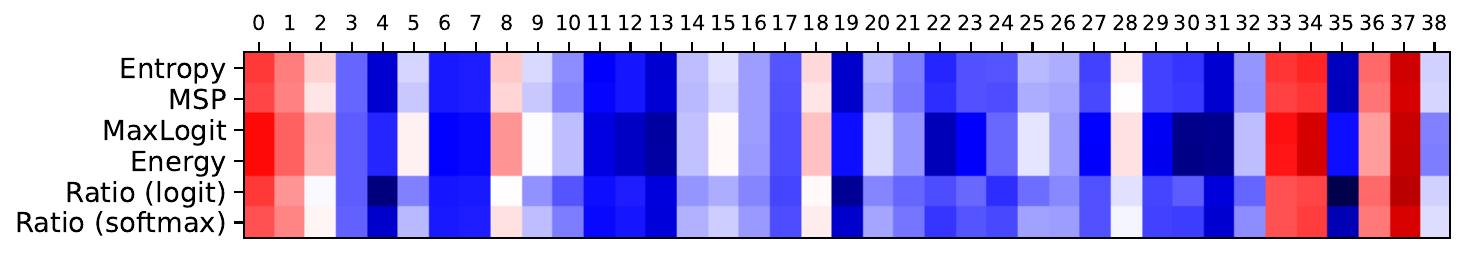}
    \caption{AUROC change of DNA quantile re-ordering compared to OOD baseline (rows) for each model (columns). Blue is better, red worse.}
    \label{fig:diff}
\end{figure}

Fig.~\ref{fig:diff} illustrates the class-specific percentual difference between AUROC scores of the DNA quantile method and different baselines for all of the 39 models/outlier classes. It can be observed that the proposed approach improves AUROC from the baseline for a majority of classes.
The differences between different OOD baselines are relatively small.
The class where DNA ordering had the largest improvement to the baseline (31~/~\textit{Polycentropus irroratus}) has an inlier class (30~/~\textit{Polycentropus flavomaculatus}), in the same genus to which 95\% of outliers are classified as.
This inlier is also the closest by DNA proximity, resulting in good OOD detection performance for the proposed approaches.
Some classes perform worse than the baseline, for example class 37~/~\textit{Sphaerium}. 41\% of the images are originally classified as \textit{Oxyethira}, which is the second furthest taxon from the outlier taxon by DNA proximity.

\subsubsection{Other results}

\begin{figure}
    \centering
    \includegraphics[scale=0.48]{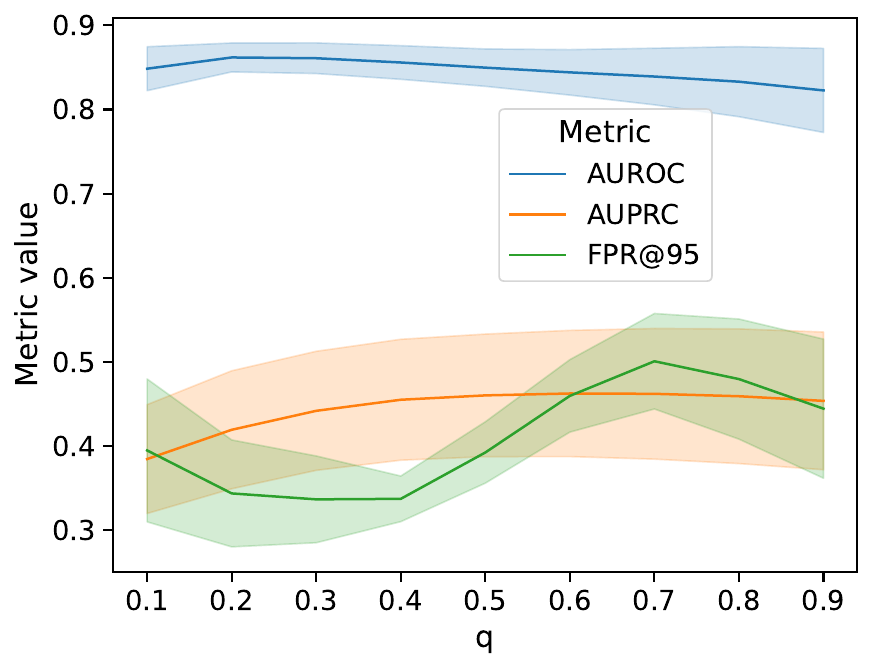}
    \caption{DNA quantile sensitivity for parameter $q$ across 39 classes for different metrics. Light area represents the standard deviation. Values were calculated with DNA distance K80 and MaxLogit OOD scoring. For AUROC and AUPRC higher is better, for FPR@95 lower is better.}
    \label{fig:q_value}
\end{figure}

The effect of the DNA quantile parameter $q$ can be seen in Fig.~\ref{fig:q_value}.
The $q$ value of $0.4$ seems to have best performance with this dataset.

\begin{figure}
    \centering
    \includegraphics[scale=0.48]{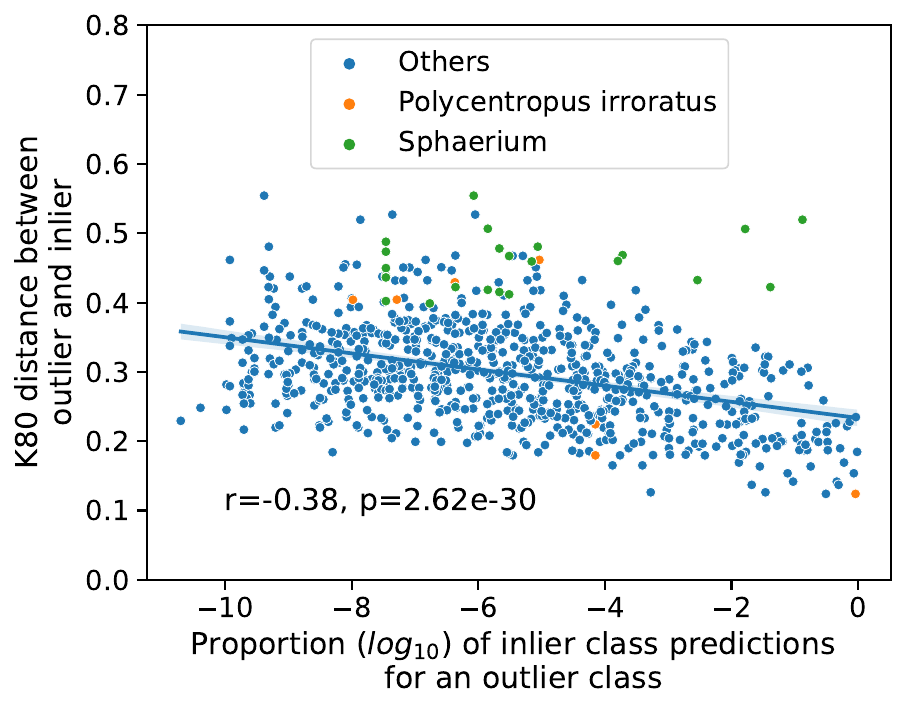}
    \caption{Comparison of the proportion of (false) inlier predictions and DNA-based distance between classes for outlier-inlier class pairs. There is a slight correlation. The outliers where re-ordering works the best (\textit{Polycentropus irroratus}) and worst (\textit{Sphaerium}) are highlighted.}
    \label{fig:proportion_distance}
\end{figure}

The classifier outputs a class prediction $c$ for each outlier image $x_{out}$.
Each prediction is an inlier: $c \in \mathcal{C}_{in}$.
Summing the predictions for all outlier images of a certain class, we can form a histogram of predictions for the outlier class, where $n_i$ is the number of outlier images predicted to inlier class $c_i$.
We can calculate the proportion of predictions in each inlier class $c_i$ as  $r_i = n_i \slash N_{out}$, where $N_{out}$ is the number of images in the outlier class.
If we do this separately for each time one of the 39 classes is an outlier and other 38 classes are inliers, we get a non-symmetric $39 \times 39$ distance matrix we can compare to the pairwise DNA distance matrix calculated for all $39\times39$ class pairs.
Omitting pairs where $r_i$ is zero and plotting the corresponding DNA distance and prediction proportion for each outlier-inlier class pair, we get the plot in Fig.~\ref{fig:proportion_distance}.
There is a slight correlation, with a Pearson R value of $-0.38$ and p value of $2.62e^{-30}$.
The figure highlights the classes from Fig.~\ref{fig:diff}, where DNA ordering had the highest improvement.

\section{Discussion}

Our results show that using DNA proximity to re-order rankings produced by common out-of-distribution detection scoring methods improves the reliability of taxonomic OOD detection, measured with several metrics.
The proposed method can be easily applied on top of any pre-trained model.

The proposed method works the best if outliers are visually classified to inlier classes that are close by DNA proximity.
This is often the case, thanks to the correlation between DNA distance and visual similarity (illustrated in Fig.~\ref{fig:proportion_distance}), especially for genus-level outliers.
However, if visually similar classes are by chance far away from the outlier by DNA distance, the method works worse than the baselines.

Methods that rely on fine-tuning the classifier with OOD samples is a common family of OOD detection methods.
DNA side information might also be useful with these methods, such when training a separate discriminator network\cite{wuhrl2024Entomoscope}, or an estimation branch \cite{devries2018Learning}.
Currently, the proposed method is limited by the assumption of a single outlier class.
In real life, DNA barcoding a group of specimens could contain several outlier classes.
This limitation will be addressed in future studies.
Datasets with realistic DNA-image pairs, such as the BIOSCAN-1M \cite{gharaee2024Step}, can be used to test the proposed methods further.

\bibliographystyle{IEEEtran}
\bibliography{IEEEabrv,bibliography}
\vspace{12pt}

\end{document}